\documentclass{article}

 \usepackage[final]{bdl_2018}
 \usepackage{seb}

\usepackage[utf8]{inputenc} 
\usepackage[T1]{fontenc}    

\title{A Unifying Bayesian View of Continual Learning}

\author{
  Sebastian Farquhar\\\
  OATML Research Group \\
  Department of Computer Science\\
  University of Oxford\\
  \texttt{sebastian.farquhar@cs.ox.ac.uk} \\
  \And
  Yarin Gal \\
  OATML Research Group \\
  Department of Computer Science\\
  University of Oxford\\
  \texttt{yarin@cs.ox.ac.uk} \\
}

\begin{document}

\maketitle

\section{Introduction}
Some machine learning applications require continual learning---where data comes in a sequence of datasets, each is used for training and then permanently discarded. Continual learning might be needed because retaining personal data is illegal or undesireable, for example a hospital may not want to keep patient data indefinitely. Alternatively, a real-time system might have so much data that complete retraining is impractical, for example, a quadcopter reacting to changing conditions. 
From a Bayesian perspective, continual learning seems straightforward: Given the model posterior one would simply use this as the prior for the next task \citep{Nguyen2018}.

However, exact posterior evaluation is intractable with many models, especially with Bayesian neural networks (BNNs) \citep{MacKay1992, Neal1995}.
Instead, posterior \textit{approximations} are often sought. In this \textit{approximate Bayesian} continual learning set up, when training on the first dataset we use a simple distribution over neural network weights to approximates the posterior given that data. 
This approximate posterior is then used as a \textit{prior} when evaluating the posterior approximation with the second dataset \citep{Nguyen2018}. Many continual learning solutions that do not rely on BNNs are justified similarly \citep{Kirkpatrick2017, Zenke2017, Chaudhry2018, Ritter2018}. We call these approaches \emph{prior-focused}. Unfortunately, when posterior \textit{approximations} are used, this prior-focused approach does not succeed in evaluations designed to capture properties of realistic continual learning use cases \citep{Farquhar2018a}.

As an alternative to prior-focused methods, we introduce a new approximate Bayesian derivation of the continual learning loss. Our loss does not rely on the posterior from earlier tasks, and instead adapts the model itself by changing the likelihood term. We call these approaches \emph{likelihood-focused}. We then combine prior- and likelihood-focused methods into one objective, tying the two views together under a single unifying framework of approximate Bayesian continual learning. This unifying view encompasses hybrid approaches which have previously not been explicitly studied.

Finally, we empirically evaluate the importance of the prior in approximate Bayesian continual learning by comparing prior-focused, likelihood-focused, and hybrid approaches. To support our analysis, we introduce Variational Generative Replay (VGR): a variational inference generalization of Deep Generative Replay (DGR) \citep{Shin2017} which falls under our likelihood-focused approaches. This model can be seen as a likelihood-focused counterpart to the current state-of-the-art prior-focused approach. We show that the likelihood-focused component within a state-of-the-art hybrid approach is responsible for all of its performance on a key benchmark. We also show that the prior-focused approach does not express well-calibrated uncertainty about whether data has been seen before, while a likelihood-focused method does. This suggests that the miscalibrated uncertainty of each successive posterior may explain why successful prior-focused approaches, despite their attractive theoretical properties, fail in some continual learning settings.

\section{Prior-focused Continual Learning}
We consider variational inference with BNNs \citep{Jordan1999, Hinton1993}. In non-continual learning, one aims to approximate a parameter posterior $p(\boldsymbol{\omega}|\mathcal{D})$ given an independently and identically distributed (i.i.d.) labelled training dataset $\mathcal{D} \equiv \{(\mathbf{x}^{(i)}, y^{(i)})\}$. In the continual learning setting, members of $\mathcal{D}$ are not i.i.d. Instead we have $T$ disjoint subsets $\mathcal{D}_{t} \equiv \{(\mathbf{x}_t^{(i)}, y_t^{(i)})\}$ each of which is individually i.i.d. and represent a task.

Variational Continual Learning (VCL) explicitly performs variational inference treating the approximate posterior distribution $q(\boldsymbol{\omega}_{t-1}|\mathcal{D}_{1:t-1})$ at the end of the $t-1$'th task as the prior when learning the $t$'th task. This entails the variational evidence lower bound loss function for the $t$'th task:

\begin{equation}
    \mathcal{L}^{t}_{VCL}(q_t(\boldsymbol{\omega})) = \sum_{n=1}^{N_t}\E_{\boldsymbol{\omega}\sim q_t(\boldsymbol{\omega})}\Big[\mathrm{log}p(y_t^{(n)}|\boldsymbol{\omega}, \mathbf{x}_t^{(n)})\Big] - \KLdiv{q_t(\boldsymbol{\omega})}{q_{t-1}(\boldsymbol{\omega})}.
    \label{eq:vcl}
    \vspace{1mm}
\end{equation}
The first term in equation (\ref{eq:vcl}) is the expected log-likelihood of the current model over the data in the current dataset. The second term penalizes difference between the current model and the approximate posterior of the previous task. Although they are not formulated using BNNs, Elastic Weight Consolidation (EWC) \citep{Kirkpatrick2017}, Synaptic Intelligence (SI) \citep{Zenke2017}, Riemanian Walk \citep{Chaudhry2018}, and the extension of EWC offered by \citet{Ritter2018} all rely on a related Bayesian justification and can be expressed with similar functional form. We call these \emph{prior-focused} methods because they treat old posteriors as the prior for a new model. These broadly correspond to what \citet{Parisi2018} call regularization approaches.

\section{Likelihood-focused Continual Learning}
Prior-focused approaches assume that $q_{t-1}(\boldsymbol{\omega})$ approximates the posterior well enough to be used for continual learning. However, we do not need to make this assumption. For example, Deep Generative Replay (DGR) \citep{Shin2017} uses successive datasets to train a Generative Adversarial Network (GAN) to simulate old datasets. These approaches build on rehearsal by \citet{Ratcliff1990ConnectionistFunctions} and the extension by \citet{Robins1995} using pseudo-rehearsal---randomly generated synthetic data.

All of these can be seen as attempts to estimate the log-likelihood of old data according to the current model, as we show below. As a result, we can call these \textit{likelihood-based} approaches. In order to capture this behaviour in a Bayesian setting, we consider the standard ELBO loss, expanded to account for the multiple datasets:
\begin{equation}
    \mathcal{L}^{t}_{ELBO}(q_t(\boldsymbol{\omega})) = \sum_{t'=1}^{t} \bigg[\sum_{n=1}^{N_t}\E_{\boldsymbol{\omega}\sim q_{t'}(\boldsymbol{\omega})}\Big[\mathrm{log}p(y_{t'}^{(n)}|\boldsymbol{\omega}, \mathbf{x}_{t'}^{(n)})\Big]\bigg] - \KLdiv{q_t(\boldsymbol{\omega})}{p(\boldsymbol{\omega})}
\end{equation}
in which the KL-divergence is always computed to the original prior and the log-likelihood is summed over all previously seen tasks. In continual learning, however, we do not have $\mathcal{D}_{t'}$ for $t' < t$ so we cannot calculate the log-likelihood directly. Instead, we estimate the expectation of the log-likelihood for each $t' < t$ separately as:

\begin{equation}\label{eq:3}
\sum_{n=1}^{N_t}\E_{\boldsymbol{\omega}\sim q_{t'}(\boldsymbol{\omega})}\Big[\mathrm{log}p(y_{t'}^{(n)}|\boldsymbol{\omega}, \mathbf{x}_{t'}^{(n)})\Big]\bigg] \approx \int \mathrm{log} \big[p(y |\boldsymbol{\omega}, \mathbf{x})\big]p_{t'}(\mathbf{x}, y) q(\boldsymbol{\omega}) d\mathbf{x}dyd\boldsymbol{\omega}.
\end{equation}

This relies on a separately learned generative model $p_{t'}(\mathbf{x}, y)$. We sample $\hat{\mathbf{x}}, \hat{y} \sim p_{t'}(\mathbf{x}, y)$ before training each task. We then create $\tilde{\mathcal{D}}_t \equiv (\tilde{\mathbf{x}}, \tilde{y}) = (\mathbf{\hat{x}} \cup \mathbf{x}, \hat{y} \cup y)$ and train on $\tilde{\mathcal{D}}_t$. Samples from a standard ELBO loss on minibatches drawn from this dataset approximate the true loss.

The joint probability distribution $p(\mathbf{x}, y)$ can be estimated in a number of ways. Early dual-memory methods are very simple approximations. Storing a small coreset or sample of the old data, as  \citet{Ratcliff1990ConnectionistFunctions}, offers a Monte Carlo approximation of the distribution. Pseudorehearsal effectively approximates $p(\mathbf{x}, y)$ by learning $p(y| \mathbf{x})$ and approximating $p(\mathbf{x}) \sim Bern(0.5)$.

\subsection{Variational Generative Replay}
 To study the contributions of the different components of Bayesian continual learning we suggest a new continual learning approach complementary to VCL. At the end of training the $t$'th task, we train a GAN using the task data to estimate $p_{t}(\mathbf{x}, y)$ and store this GAN. The collection of stored GANs are sampled to create $\tilde{\mathcal{D}}_{t+1}$. We call this Variational Generative Replay (VGR). DGR can be seen as a maximum likelihood estimation approximation of our approach, with some other implementation differences.

\section{A Unifying View of Continual Learning}
We can combine prior- and likelihood-focused approaches, drawing on the strengths of each. We use the approximation of (\ref{eq:3}) alongside a fresh prior for each task to form a hybrid loss function:
\begin{equation}
    \mathcal{L}^{t}_{Hybrid}(q_t(\boldsymbol{\omega})) = \sum_{t'=1}^{t} \bigg[\int \mathrm{log} \big[p(y |\boldsymbol{\omega}, \mathbf{x})\big]p_{t'}(\mathbf{x}, y) q(\boldsymbol{\omega}) d\mathbf{x}dyd\boldsymbol{\omega}\bigg] - \KLdiv{q_t(\boldsymbol{\omega})}{q_{t-1}(\boldsymbol{\omega})}.
    \label{eq:hybrid}
\end{equation}

For a hybrid approach, part of each dataset is used to estimate a posterior for future tasks, while some is reserved to estimate a generative model of old data. Although they do not explicitly interpret it as such, two hybrid approaches have already been presented. \citet{Nguyen2018} extend VCL with coresets---small held-out samples from earlier datasets---which are used to fine-tune the model at the end of each task. This approximates a special case of (\ref{eq:hybrid}), though the generative and current components of the likelihood are trained separately, rather than using a single loss function as we suggest. \citet{Chaudhry2018} also implicitly approximate (\ref{eq:hybrid}) when they supplement training with a sample of data from old tasks.

\begin{table}
    \centering
    \caption{Comparison of methods evaluated in this abstract}
    \vspace{-2mm}
    \begin{tabular}{rcc}
        \toprule
         & \textbf{Prior} & \textbf{Likelihood given...}\\
        \midrule
         VGR & $p(\boldsymbol{\omega})$ & Generative models of old tasks\\
         VCL & $q_{t-1}(\boldsymbol{\omega})$ & Current task\\
         VCL w/ coresets & $q_{t-1}(\boldsymbol{\omega})$ & Current task \& Coresets \\
         Coreset only & $p(\boldsymbol{\omega})$ & Current task \& Coresets\\
         \bottomrule
    \end{tabular}
    \label{tb:methods}
\end{table}

\section{Experimental Analysis}
\begin{figure}
    \begin{minipage}{.45\textwidth}
        \includegraphics[width=\linewidth]{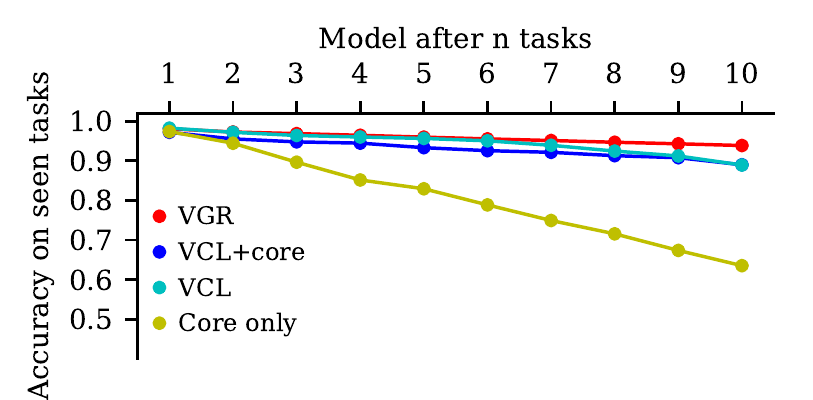}
        \caption{Permuted MNIST. All main methods perform well, although coresets on their own do not.}
        \label{fig:permuted_all}
        \vspace{4mm}
    \end{minipage}
    \hfill
    \begin{minipage}{.45\textwidth}
        \includegraphics[width=\columnwidth]{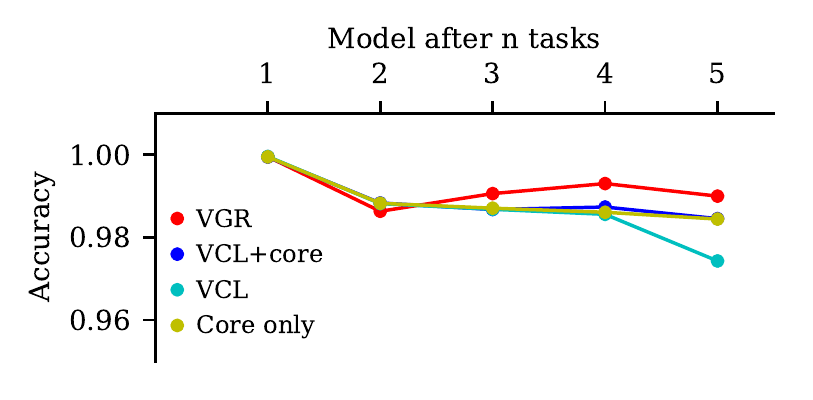}
        \caption{Multi-headed Split MNIST. All methods perform well.}
        \label{fig:split_multi_all}
        \vspace{8mm}
    \end{minipage}
    \centering
    \begin{minipage}{.45\textwidth}
        \vspace{-5mm}
        \includegraphics[width=\columnwidth]{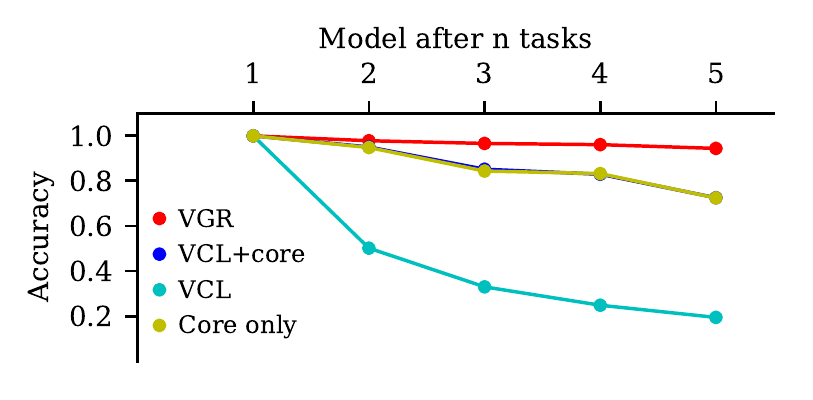}
        \vspace{-5mm}
        \caption{VGR performs well on single-headed Split MNIST. VCL performance is determined by coreset.}
        \label{fig:split_single_all}
    \end{minipage}
\end{figure}
We empirically assess the importance of the prior in prior-focused and hybrid continual learning by isolating the different components of VCL and comparing their performance with VGR.\footnote{We do not compare VCL with DGR directly, as the use of mean-field variational inference in the former but not the latter makes it hard to isolate the difference between likelihood- and prior-focused continual learning.}  We compare the four variant models shown in Table \ref{tb:methods}, using implementations by \citet{Nguyen2018} for VCL. Experimental details can be found in Appendix \ref{details}.

VCL (with and without coresets) and VGR perform well on simple tasks like Permuted MNIST (as described in \citet{Goodfellow2013}, see fig \ref{fig:permuted_all}) and multi-headed Split MNIST (as described by \citet{Zenke2017}, see fig \ref{fig:split_multi_all}). We present these as standard baseline experiments, despite the significant shortcomings shown in \citet{Farquhar2018a}. These show that likelihood- and prior-focused approaches as well as hybrid ones have good performance on simple experiments.

However, when Split MNIST is performed without multi-heading (see \citet{Chaudhry2018, Farquhar2018a}), we see that VGR significantly outperforms VCL (figure \ref{fig:split_single_all}). Moreover, we show by ablation that coresets (the likelihood-focused component) contribute all of the performance of the hybrid approach, since the coresets only model performs exactly as well as VCL with coresets. This indicates that the posterior approximations used for continual learning are insufficient.

\begin{figure}[b]
    \begin{minipage}{.45\textwidth}
        \includegraphics[width=\columnwidth]{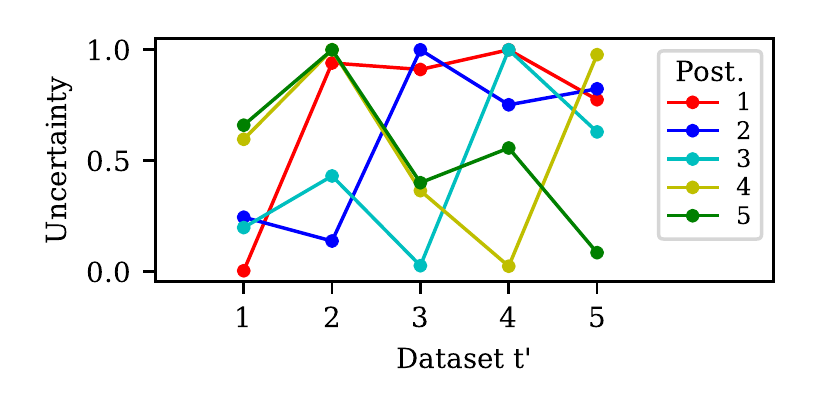}
        \vspace{-4mm}
        \caption{Each posterior's scaled uncertainty is shown for VCL. It should be low for only datasets with $t' \leq t$ but is not.}
        \label{fig:mi_VCL}
    \end{minipage}
    \hfill
    \begin{minipage}{.45\textwidth}
        \includegraphics[width=\columnwidth]{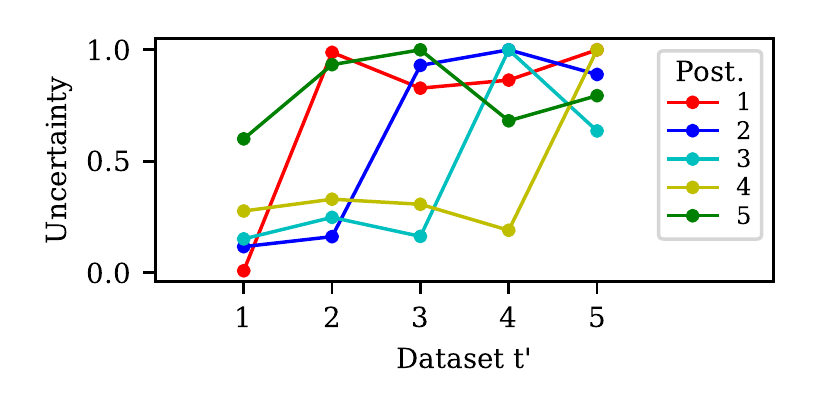}
        \vspace{-4mm}
        \caption{For VGR uncertainty (rescaled mutual information) is low on seen tasks and high on unseen ones as it should be.}
        \label{fig:mi_VGR}
    \end{minipage}
\end{figure}

When we specifically examine the posteriors used by VCL, we see that the uncertainty of the prior model does not distinguish seen from unseen datasets after the first task (figure \ref{fig:mi_VCL}). We assess uncertainty using the mutual information between the model parameters and data labels given the test dataset for each of the five datasets for each of five models (shown scaled by the largest mutual information for each model). However, the uncertainty measured for the posteriors from VGR (which includes training on generated data) performs much better (see fig \ref{fig:mi_VGR}). The posteriors from VGR clearly distinguish seen from unseen data.

\section{Discussion}
Although prior-focused methods have a Bayesian justification, they assume that model posteriors are good enough to be used as priors for learning. We show that this assumption can be too strong, evidenced by poor performance on single-headed Split MNIST and the fact that VCL models do not have well calibrated uncertainty about unseen data. Instead, we show that existing dual-memory approaches can be interpreted from a Bayesian perspective. We show that simple likelihood-focused methods can outperform prior-focused methods and can offer better uncertainty estimates.

\newpage
\bibliography{references}
\bibliographystyle{plainnat}

\newpage

\appendix
\section{Experimental Details}\label{details}

For permuted MNIST, we follow \citet{Nguyen2018} where possible using an implementation made public by those authors. We use a Bayesian neural network \citep{Graves2011} with two hidden layers of 100 units with ReLu activations. The priors are initialized as a unit Gaussian and the parameters are initialized with the mean of a pre-trained maximum likelihood model and a small initial variance ($10^{-6}$). We use the Adam optimizer \citep{Kingma2015} with learning rate $10^{-3}$. For both VGR and VCL we use a single head, again following \citet{Nguyen2018}. For all results we present the average over 10 runs, with a different permutation each time. Standard errors are not shown as they are under a tenth of a percent.

For VCL we train for 100 epochs using a batch size of 256 on all the data except the with-held coresets. We train the whole single head on coresets for 100 epochs and use 200 digits of each permutation as a coreset chosen using the same k-center coresets algorithm used by \citet{Nguyen2018}. VCL without coresets is exactly the same, but without a final training step on coresets. The coresets only algorithm is exactly the same as VCL, except that the prior is always initialized as though it were the first task.

We train VGR for 120 epochs using a GAN trained for 200 epochs on each MNIST digit, using the same data which is used for the main training process. We use 6000 generated digits per class, sampled fresh for each task, and initialize network weights using the previous task. We use a batch size of 256 times the number of seen tasks, ensuring that the number of batches is held constant.

The GAN is trained with an Adam optimizer with learning rate $2*10^{-4}$ and $\beta_1$ of $0.5$. The network has four fully-connected hidden layers with 256, 512, 1024 and 784 weights respectively. It uses Leaky ReLu with $\alpha$ of 0.2.

The settings for Split MNIST follow \citet{Nguyen2018} where possible. We use exactly the same architecture as for Permuted MNIST, including the single head, except that each hidden layer has 256 weights, similarly to \citet{Nguyen2018} on their multi-headed Split MNIST. Results are shown averaged over 10 runs, with a different coreset selection each time. Standard errors are not shown as they are of the order of a tenth of a percent.

For all architectures we train for 120 epochs. For VCL we use batch sizes equal to the training set size. We use coresets of 40 digits per task selected using the same k-center coreset algorithm as \citet{Nguyen2018}, which are withheld from the training. We train for 120 epochs on the concatenated coreset across all the heads together.

For VGR, we use 6000 digits per class generated by a convolutional GAN. Unlike VCL, we cap batch sizes at 30,000 rather than having the batch size equal the training set size. This is because the examples generated by VGR result in larger training sets which exceeded the memory available on our GPU.The GAN is trained for 50 epochs on each MNIST class using the same optimizer as the non-convolutional GAN used for Permuted MNIST. It has a fully connected layer followed by two convolutional layers with 64 and 1 channel(s) and 5x5 convolutions. Each convolutional layer is preceded by a 2x2 up-sampling layer. The activations are Leaky ReLu's with $\alpha$ of $0.2$.

Multi-headed Split MNIST has almost precisely the same settings as the Single-headed Split MNIST above. Following \citet{Nguyen2018}, for VCL we have five heads in the final layer, each of which is trained separately. Coresets are used to train each head in turn. Batch size equal is to training set size. Results are shown averaged over 10 runs with fresh coreset selection each time. Standard errors are not shown as they are of the order of a tenth of a percent.

For VGR, unlike VCL, we only perform multi-heading as a way of selecting predictions, rather than using multiple heads during training as well, because each batch contains data from multiple tasks.
\end{document}